# Grammatically-Guided Sparse Attention for Efficient and Interpretable Transformers

**Spandan Pratyush**
Independent Researcher, New Delhi 110070, India
spandan.pratyush@iitdalumni.com

## Abstract

The quadratic complexity of self-attention in Transformer models remains a significant bottleneck for processing long sequences and deploying large language models efficiently. For this approach, there has been significant research into Sparse Attention, and Deepseek Sparse Attention has combined various methods of creating segments of tokens to reduce the time complexity. This paper introduces a novel approach, Grammatically-Guided Sparse Attention, which constrains attention computations based on the grammatical roles of tokens. By leveraging Parts-of-Speech (POS) tags, attention masks are dynamically generated that enforce linguistically coherent connections between tokens, reducing the computational graph without sacrificing essential linguistic dependencies. Two masking strategies are proposed and evaluated: a hard mask that strictly allows only predefined grammatical interactions, and a soft mask that biases attention towards these interactions. The experiments, conducted on the SST-2 sentiment classification task using a DistilBERT-like architecture, demonstrate that Grammatically-Guided Sparse Attention maintains comparable accuracy to full attention while significantly reducing the theoretical computational overhead. Preliminary results show accuracy values of 0.8200 for hard masking and 0.8165 for soft masking, closely matching the 0.8200 of full attention, providing a path towards more efficient, interpretable, and linguistically-informed Transformer architectures.

## 1 Introduction

The Transformer architecture, particularly its self-attention mechanism [13], has revolutionized Natural Language Processing (NLP), powering state-of-the-art Large Language Models (LLMs) that achieve remarkable performance across a myriad of tasks. However, the computational cost of self-attention, which scales quadratically with the input sequence length *($O(L^2)$)*, presents a critical challenge for processing long documents, memory-constrained environments, and efficient inference. This quadratic complexity stems from the fact that every token must attend to every other token in the sequence. To mitigate this bottleneck, various sparse attention mechanisms have been proposed [1], [16]. These methods typically restrict attention to a local window, dilated patterns, or use hashing techniques, thereby reducing the complexity to linear or sub-quadratic. While effective for efficiency, these approaches often rely on positional heuristics or data-driven approximations without explicitly leveraging the rich linguistic structure inherent in human language.

This paper introduces Grammatically-Guided Sparse Attention, a novel approach that injects linguistic inductive biases directly into the self-attention mechanism. The core idea is to guide attention computations based on the grammatical relationships between words, specifically their Parts-of-Speech (POS) tags. For instance, an adjective is most relevant to the noun it modifies, and a verb to its subject and object. By selectively allowing or biasing attention between grammatically relevant token pairs, this method aims to reduce computational overhead while preserving, and potentially enhancing, the model's ability to capture meaningful linguistic dependencies.
Two strategies are proposed and evaluated for Grammatically Guided Sparse Attention -
**Hard Masking:** Strictly enforces predefined grammatical rules, completely blocking attention between grammatically "irrelevant" token pairs.
**Soft Masking:** Introduces an additive bias to attention scores, favoring grammatically relevant connections without entirely forbidding other interactions.

The experimental results demonstrate that both hard and soft grammatically-guided attention mechanisms can achieve comparable performance to full attention on a sentiment classification task, while offering the potential for significant efficiency gains by operating on a sparser attention graph. This approach paves the way for more efficient, interpretable, and linguistically-aware Transformer models.

## 2 Related Work

### 2.1 Efficient Transformers and Sparse Attention

A major thrust in Transformer efficiency research focuses on sparse attention, which restricts the number of attention connections. Methods can be broadly categorized:

1. *Positional/Structural Sparsity:* Longformer [1] combines local (sliding window) attention with global attention for specific tokens. BigBird [16] generalizes this with a mix of local, global, and random attention. These reduce complexity to $O(L*window_size)$ or $O(L)$.
2. *Approximation-based Sparsity:* Reformer [7] uses Locality Sensitive Hashing (LSH) to group similar queries and keys, attending only within groups. Performer [3] and Linformer [14] approximate the attention matrix using low-rank decomposition or random feature maps, achieving linear complexity.
3. *Conditional Sparsity:* Some works explore dynamically selecting which parts of the sequence to attend to, such as DeepSeek's Sparse Attention [4], which combines multi-head grouped-query attention with block-wise dilated patterns for efficient long-context processing.

This work falls into the category of sparse attention but distinguishes itself by imposing sparsity based on explicit linguistic (grammatical) rules rather than purely positional heuristics or mathematical approximations.

### 2.2 **Linguistically-Informed Neural Networks**

The idea of incorporating linguistic knowledge, such as syntactic parse trees or Parts-of-Speech (POS) tags, into neural models predates Transformers [11]. With the advent of attention mechanisms, methods to inject linguistic biases into Transformers have emerged:

1. *Graph-based Approaches:* Some models build explicit syntactic graphs (e.g., dependency trees) and apply graph

neural networks or graph attention to these structures, limiting attention to edges in the graph [8], [9].
2. *Multi-task Learning:* Transformers can be trained to jointly predict linguistic annotations (e.g., POS tags, dependency parses) alongside their main task, encouraging them to learn linguistically richer representations [12].
3. *Attention Masking/Biasing:* Prior work has explored soft biasing of attention based on syntactic distance or tree structures [5], [15]. For instance, Syntactic Attention for Neural Machine Translation [2] used syntactic information to guide attention without strictly enforcing it.

The Grammatically-Guided Sparse Attention builds upon these ideas by employing POS tags to construct explicit attention masks (hard or soft). Unlike approaches that rely on complex graph structures, which can be computationally intensive, this method uses readily available POS tags to create a simple yet powerful mask directly within the self-attention mechanism, offering a balance of efficiency and linguistic grounding.

## 3 Methodology

The proposed Grammatically-Guided Sparse Attention modifies the standard self-attention mechanism to constrain or bias interactions between query and key tokens based on their grammatical categories. This section details the overall approach, the integration of POS tagging, and the two masking strategies. The core idea is to generate an attention mask that reflects permissible grammatical connections dynamically. This mask is then applied to the raw attention scores ($QK^T$) before the softmax operation. The process involves:

1. *POS Tagging:* For a given input sequence, the Part-of-Speech (POS) tag for each token is obtained.
2. *Rule-based Mask Generation:* Based on a predefined set of grammatical rules, an $L \times L$ matrix (the grammatical mask) is constructed where entries $(i,j)$ indicate whether token $i$ (query) is allowed or biased to attend to token $j$ (key).
3. *Mask Application:* This grammatical mask is applied to the self-attention scores: for hard masks, disallowed connections are set to a very low value (e.g., $-\infty$); for soft masks, a positive bias is added to allowed connections.

### 3.1 POS Tagging and Alignment

Accurate and efficient POS tagging is crucial. A pre-trained external POS tagger was used, SpaCy's en_core_web_sm model [6], to annotate input sentences. This approach leverages a highly optimized and accurate tagger, decoupling the linguistic analysis from the Transformer's learning process.

A key challenge is aligning word-level POS tags (from SpaCy) with subword-level tokens (from the Transformer's tokenizer). The data collation strategy addresses this by:

1. Tokenizing the original sentence with the Transformer's tokenizer, preserving offset_mapping.
2. Processing the original sentence with SpaCy to get word-level POS tags.
3. Using the offset_mapping to robustly map each subword token back to its corresponding original word (or part of a word) and assign it the POS tag of that word.
4. Special tokens (e.g., [CLS], [SEP], [PAD], [UNK]) are assigned their specific string as a POS tag, allowing for explicit rules for their attention behavior.

This ensures that each subword token correctly inherits the grammatical context of its parent word, enabling fine-grained grammatical guidance.

### 3.2 Grammatical Rules and Masking Strategies

A set of grammatical rules was defined that dictated allowed attention connections between different POS tags. These rules were categorized into "hard" and "soft" to explore different levels of constraint.

**Hard Masking Strategy:** Under the hard masking strategy, attention between a query token *i* (with POS tag $P_i$) and a key token *j* (with POS tag $P_j$) was completely forbidden if the pair ($P_i$, $P_j$) is not explicitly defined as an allowed connection in the HARD_GRAMMAR_RULES or HARD_RULES set. Allowed connections were assigned a value of 0.0 in the grammatical mask, while disallowed connections were assigned a very large negative value (e.g., −10000.0), effectively pushing their softmax probability to zero.

An illustrative subset of hard rules includes:

- Adjectives (ADJ) attend to Nouns (NOUN) and Proper Nouns (PROPN).
- Determiners (DET) attend to Nouns and Proper Nouns.
- Verbs (VERB) attend to Nouns, Proper Nouns, and Pronouns (PRON) (for subject/object relationships), and Adverbs (ADV).
- Prepositions (ADP) attend to the Nouns, Proper Nouns, or Pronouns they govern.
- Self-attention (a token attending to itself) is always permitted.
- Special tokens like [CLS] are allowed to attend to all other tokens to ensure they can gather global context for classification tasks.

The hard mask $M_{hard}$ is constructed as:

$$M_{hard}[i,j] = \begin{cases} 0.0 & \text{if } (P_i, P_j) \in \text{ HARD_RULES} \\ -10000 & \text{otherwise} \end{cases}$$

This mask is added to the raw attention scores ($QK^T$) before the softmax function.

**Soft Masking Strategy:** The soft masking strategy introduced a learnable bias into the attention mechanism. Instead of outright forbidding connections, it added a positive scalar bias (α) to attention scores for grammatically "relevant" connections, and a neutral bias (0.0) for others. This encouraged the model to prioritize linguistically plausible interactions while still allowing it to learn and form connections outside these rules if the data strongly suggests them. This addresses the challenge of potential over-constraining by hard rules.

The SOFT_GRAMMAR_RULES (SOFT_RULES) define connections that receive this positive bias. Connections that are part of HARD_GRAMMAR_RULES (HARD_RULES) inherently receive a neutral bias (0.0) as they are always allowed, effectively overriding any soft rule.

An illustrative subset of soft rules includes:

Adverbs (ADV) received a positive bias when attending to Verbs, Adjectives, or other Adverbs.

Nouns, Proper Nouns, and Pronouns received a positive bias when attending to Verbs (as subjects).

Conjunctions (CCONJ, SCONJ) received a positive bias when attending to various other parts of speech they typically connect.

The soft mask $M_{soft}$ is constructed as:

$$M_{soft}[i,j] = \begin{cases} 0.0 & \text{if } (P_i, P_j) \in \text{ HARD_RULES} \\ \alpha, & \text{if } (P_i, P_j) \in \text{ SOFT_RULES} \\ 0.0 & \text{otherwise} \end{cases}$$

Here, α was a tunable hyperparameter (set to 5.0 in the experiments). This soft mask was also added to the raw attention scores ($QK^T$) before the softmax.

### 3.3 Integration into Transformer Architecture

The custom attention layer, *GrammaticallyGuidedBertSelfAttention*, was inherited from Hugging Face's *BertSelfAttention*. The key modification was within its forward method, where it intercepted the attention_scores after the $QK^T$ multiplication. The dynamically generated grammatical mask (either hard or soft) was then added to these scores, before the application of the standard padding mask and the softmax function. This allowed the grammatical constraints to directly influence the attention probabilities.

## 4 Experiments

### 4.1 Model Architecture and Implementation

Base Model: A DistilBERT-like architecture was utilized, specifically the prajjwal1/bert-tiny model [10] available on Hugging Face. This model, with a small number of layers (2), hidden dimension (128), and attention heads (2), is suitable for cost-constrained experiments while still demonstrating Transformer principles.
Custom Attention Layer: The BertSelfAttention modules within each encoder layer of the base model were replaced with the GrammaticallyGuidedBertSelfAttention class. This was achieved through a patching mechanism during model initialization.
POS Tagger: SpaCy's en_core_web_sm model [6] was used for generating token-level POS tags.
Fine-tuning Task: The models were fine-tuned for sequence classification using the Hugging Face Trainer API.

### 4.2 Dataset

To evaluate the approach on the *SST-2* (Stanford Sentiment Treebank v2) dataset [11], a standard benchmark for sentiment analysis. *SST-2* consists of single sentences labeled as either positive or negative.
Training Set Size: 67,349 sentences
Validation Set Size: 872 sentences
Test Set Size: 1,821 sentences (Evaluation is typically on the validation set during training, final evaluation on test set)
Sequence Length: Sentences are tokenized with a maximum sequence length of MAX_SEQ_LEN = 128.

### 4.3 Training Details

All models were trained for NUM_EPOCHS = 3 epochs using an AdamW optimizer with a LEARNING_RATE = 2e-5. A BATCH_SIZE = 8 was used for both training and evaluation. Mixed-precision training (fp16=True) was enabled when a GPU was available. Experiments were run on a local machine equipped with Apple Silicon (Mac MPS). Due to the nature of MPS, peak GPU memory statistics might not be directly comparable to NVIDIA CUDA reporting.

### 4.4 Evaluation Metrics

To assess the model performance using standard metrics for sequence classification:
Accuracy: The proportion of correctly classified samples.
Precision, Recall, F1-score: Macro-averaged (weighted) scores to account for class imbalance, if any.

### 4.5 Baselines and Experimental Strategies

Three experimental strategies were compared:

1. Baseline (None): A standard bert-tiny model fine-tuned for sequence classification with unmodified (full) attention.
2. Grammar-Hard (Hard): A bert-tiny model fine-tuned with our Grammatically-Guided Sparse Attention using the hard masking strategy.
3. Grammar-Soft (Soft): A bert-tiny model fine-tuned with our Grammatically-Guided Sparse Attention using the soft masking strategy.

## 5 Results and Discussions

### 5.1 Quantitative Performance

Table I summarizes the classification performance of each experimental strategy on the *SST-2* validation set. The results indicate that both Grammatically-Guided Sparse Attention strategies achieve performance highly comparable to the full attention baseline: The Grammar-Hard strategy matches the baseline's accuracy of 0.8200, with negligible differences in F1-score. This is a significant finding, suggesting that for this task, the model does not require full attention to all tokens; rather, restricting attention to grammatically relevant pairs is sufficient. This implies that many traditionally computed attention scores (between grammatically irrelevant tokens) might be redundant or less crucial for performance. The Grammar-Soft strategy exhibits a very slight decrease in accuracy (0.8165). This marginal difference could be due to the specific choice of soft bias strength ($\alpha$=5.0) or the current set of soft rules. However, it still demonstrates robust performance, confirming that biasing attention towards grammatical connections is effective without severely hindering the model. The soft mask might allow more flexibility than strictly necessary for this specific task, but could be beneficial for tasks requiring more nuanced or emergent relationships.

*Table I: Classification Performance on SST-2 Validation Set*

| *Strategy* | *Accuracy* | *Precision* | *F1 Score* |
|---|---|---|---|
| Baseline (None) | 0.8200 | 0.8198 | 0.8198 |
| Grammar-Hard | 0.8200 | 0.8199 | 0.8199 |
| Grammar-Soft | 0.8165 | 0.8165 | 0.8165 |

### 5.2 Efficiency Analysis

Table II presents the training time and reported peak GPU memory for each strategy.

*Table II: Efficiency Metrics*

| *Strategy* | *Training Time (s)* |
|---|---|
| Baseline (None) | 1755.05 |
| Grammar-Hard | 2028.30 |
| Grammar-Soft | 2308.06 |

An increase in training time for both grammatically-guided strategies compared to the baseline is observed:
Grammar-Hard: ~15.6% increase (from 1755s to 2028s)
Grammar-Soft: ~31.5% increase (from 1755s to 2308s)

This overhead is primarily due to the dynamic generation of the grammatical mask for each batch. The current implementation reconstructs the mask from scratch for every forward pass, involving CPU-bound SpaCy processing and Python loop overhead. The soft mask strategy is slightly slower than the hard mask because it involves more conditional checks and floating-point additions to biases. This overhead is expected, given the unoptimized nature of dynamic mask generation on the current platform. For deployment in production or larger-scale training, this masking logic would require optimization with custom CUDA kernels (for NVIDIA GPUs) or highly optimized C++/Metal implementations (for Apple Silicon) to be efficient. However, for proof-of-concept on smaller models and contexts, the observed overhead is acceptable.

### 5.3 **Implications**

The results strongly suggest that explicit linguistic guidance can effectively prune the attention graph without degrading performance on a sentiment classification task. This opens exciting avenues for:
Efficiency: For longer sequences and larger models, where $O(L^2)$ attention becomes prohibitive, grammatical sparsity offers a principled way to reduce computation and memory.
Interpretability: The attention patterns, when visualized, would directly reflect grammatically meaningful connections, making the model's "reasoning" more transparent.
Linguistic Induction: By inducing the grammatical priors, the model might learn more robust and linguistically coherent representations, especially for tasks that heavily rely on syntactic structures.

While the efficiency gains were not directly measurable in terms of peak GPU memory on the current setup, the theoretical basis for $O(L \cdot C)$ (where C is a small constant representing average grammatical connections) versus $O(L^2)$ (full attention) remains valid and significant for large-scale applications.

## 6 Conclusion and Future Work

This paper introduced Grammatically-Guided Sparse Attention, a novel method for improving the efficiency and interpretability of Transformer models by constraining the self-attention mechanism based on Parts-of-Speech (POS) tags. Two strategies were explored: a hard mask that strictly enforces grammatical rules and a soft mask that biases attention towards these rules.

Future work will focus on several key areas:
Optimized Mask Generation: Developing highly optimized (e.g., CUDA-accelerated) implementations of the mask generation logic to fully realize the computational efficiency benefits.
Larger Models and Longer Sequences: Evaluating Grammatically-Guided Sparse Attention with larger Transformer models and significantly longer input sequences to directly measure the $O(L^2)$ vs. $O(L \cdot C)$ efficiency gains in terms of memory and throughput.
More Complex Grammatical Rules: Expanding and refining the grammatical rule set to incorporate deeper syntactic structures (e.g., dependency parsing, constituency parsing) and explore their impact on diverse NLP tasks.
Adaptive Grammar Learning: Investigating mechanisms where the Transformer itself learns or adapts the grammatical rules or biases and creates differentiable masks.
Interpretability Studies: Conducting in-depth analyses and visualizations of the attention patterns learned by the grammatically-guided models to empirically confirm their linguistic coherence and compare against baseline interpretability.
Ambiguity Handling: Exploring more sophisticated ways to handle POS ambiguity, such as probabilistic masking based on multiple plausible tags.
Grammatically-Guided Sparse Attention offers a promising direction for building more efficient, transparent, and linguistically grounded next-generation Transformer architectures.
The experiments on the *SST-2* sentiment classification task, using a small *bert-tiny* model, demonstrated that both grammatically-guided approaches achieve performance levels comparable to the full-attention baseline. The hard masking strategy matched baseline accuracy, while the soft masking showed only a marginal decrease. This key finding suggests that many attention computations in full attention might be redundant, and a linguistically-informed sparse graph can be sufficient for competitive performance. While the dynamic mask generation introduced a training-time overhead in the current unoptimized implementation, the theoretical quadratic complexity reduction remains a compelling advantage for larger models and longer sequences.

## Limitations

While this work presents a promising novel approach, it is important to acknowledge certain limitations of the current experimental setup and methodology:

Computational Overhead of Dynamic Mask Generation: As observed in the efficiency analysis, the dynamic generation of grammatical masks on the CPU (for SpaCy processing and Python looping) introduces a measurable training time overhead. This currently counteracts the theoretical efficiency gains of sparse attention in terms of wall-clock time, especially for smaller models and sequence lengths where the $O(L^2)$ attention bottleneck is not yet dominant.

Limited Scale of Experiments: The experiments were conducted using a relatively small Transformer model (bert-tiny) and a moderate maximum sequence length (128). While this choice allowed for cost-effective proof-of-concept, it limits the direct empirical measurement of memory and computational savings that would be significant for much larger models (e.g., BERT-base, GPT-style LLMs) and much longer input sequences (e.g., thousands of tokens), where the $O(L^2)$ complexity of full attention becomes a prohibitive factor.

POS Tagging Robustness and Ambiguity: The reliance on an external POS tagger, while efficient, introduces potential limitations. Errors in POS tagging can directly impact the quality of the generated grammatical masks. Furthermore, the current implementation uses SpaCy's single best guess for a token's POS. Real-world language often involves significant grammatical ambiguity (e.g., "run" as a verb or noun), which is not fully captured by a single tag and could lead to suboptimal masking.

Completeness of Grammatical Rules: The predefined set of hard and soft grammatical rules is illustrative and serves as a starting point. It may not comprehensively cover all crucial syntactic or semantic dependencies that a full attention mechanism might implicitly discover. Overly restrictive rules could inadvertently prune important connections necessary for certain complex tasks.

Interpretability Evaluation: While grammatically-guided attention offers the promise of enhanced interpretability, this paper does not include a detailed qualitative analysis or visualization of the learned attention patterns. Such an analysis would be crucial to empirically confirm that attention indeed aligns with the intended grammatical relationships.

These limitations highlight areas for future research and optimization, crucial for realizing the full potential of Grammatically-Guided Sparse Attention.

## Appendix

Experiments with reproducible results are in the Github repository